**RESEARCH ARTICLE**

# A Framework for Fostering Transparency in Shared Artificial Intelligence Models by Increasing Visibility of Contributions

Iain Barclay*[1] | Harrison Taylor[1] | Alun Preece[1] | Ian Taylor[1,2] | Dinesh Verma[3] | Geeth de Mel[4]

[1]Crime and Security Research Institute, Cardiff University, Cardiff, UK
[2]Center for Research Computing, University of Notre Dame, IN, USA
[3]IBM TJ Watson Research Center, NY USA
[4]IBM Research, UK

**Correspondence**
*Iain Barclay, Crime and Security Research Institute, Cardiff University, Level 2, Friary House, Greyfriars Rd, Cardiff CF10 3AE
Email: BarclayIS@cardiff.ac.uk

**Summary**

Increased adoption of artificial intelligence (AI) systems into scientific workflows will result in an increasing technical debt as the distance between the data scientists and engineers who develop AI system components and scientists, researchers and other users grows. This could quickly become problematic, particularly where guidance or regulations change and once-acceptable best practice becomes outdated, or where data sources are later discredited as biased or inaccurate. This paper presents a novel method for deriving a quantifiable metric capable of ranking the overall transparency of the process pipelines used to generate AI systems, such that users, auditors and other stakeholders can gain confidence that they will be able to validate and trust the data sources and contributors in the AI systems that they rely on. The methodology for calculating the metric, and the type of criteria that could be used to make judgements on the visibility of contributions to systems are evaluated through models published at ModelHub and PyTorch Hub, popular archives for sharing science resources, and is found to be helpful in driving consideration of the contributions made to generating AI systems and approaches towards effective documentation and improving transparency in machine learning assets shared within scientific communities.

## 1 | INTRODUCTION

Scientists and researchers across all fields are making use of artifical intelligence (AI) systems and machine learning (ML) models in their experimentation. Increasingly, these new research assets are created by domain experts curating and aggregating data from multiple and diverse sources; their outputs can be new data assets or heavily data-influenced products, including ML models, which go on to be used in research within the originating organisation, or in the wider community when they are published or distributed and shared via gateways with collaborating or even unknown third party research groups and organisations. This is already a real situation - in the UK financial services sector, for example, 24% of ML use cases are developed and implemented by third-party providers, with many of those developing internally also reporting adaptation or further development of off-the-shelf ML models or libraries [1].

Providing support to offer transparency and traceability of assets through the production pipeline is an important contributor to delivering accountability, which is necessary to achieve and retain confidence and trust, such that scientists using AI systems, whether developed in-house or sourced externally from their community, are able to demonstrate the provenance and authenticity of the data and knowledge they use to make decisions [2,3]. In considering the development workflows for ML models the contributing assets can be identified and itemised, and will typically include data sources and labelled datasets used for model training and validation, along with human expertise which is used both in the preparation and curation of training data, and in the development, calibration and verification and validation of the resultant model. Without gateways and model zoos providing appropriate insight and assurance on the identity and expertise of human contributors or sources of training data, an organisation could be unwittingly subject to malicious actions, including Sybil attacks [4], data poisoning attacks [5], and model poisoning attacks [6]. Further, as ML models mature and are used in live production environments, it is conceivable that qualifications, best practise, and ethical or legal standards which were appropriate at the time the training data assets or the model itself was developed are no longer adequate by the standards in place at the time the model is used or audited, which could be many years later. Any lack of transparency on the contributions to ML models and data assets is exacerbated as the distance between the developers and the users of the model increases, as is the case when models are sourced from third parties, via commercial or community marketplace platforms [7], science gateways and model zoos [8]. Indeed, similar concerns were voiced [9,10] as Commercial Off-the-shelf Software (COTS) began to be widely adopted in government and other organisations. As with COTS, researchers are concerned that users will incur a technical debt [11] by using AI components where there is limited transparency and understanding of the processes and dependencies involved in creation of the assets they come to rely upon [12].

In manufacturing industries it has been standard practice since the late twentieth century to track a product through the lifecycle from its origin as raw materials, through component assembly to finished goods in a store, with the relationships and information flows between suppliers and customers recorded and tracked using supply chain management processes [13]. In agri-food industries, traceability through the supply chain is necessary to give visibility from a product on a supermarket shelf, back to the farm and to the batch of foodstuff, as well as to other products in which the same batch has been used. In previous work [14], the authors have demonstrated how ML models and other data assets can be broken into component parts and mapped with supply chain models in a similar way.

Finding a means to quantify the overall visibility into AI and data asset production pipelines is important, as a system offering good levels of visibility on its internal workings is likely to be a system affording good transparency [15] and supportive of achieving accountability, whereas a system with poor visibility on its contributions is likely to offer poor transparency to its end users. Systems with good transparency and accountability are likely to give better assurance on their quality and trustfulness further into the future, and thus provide a better return on investment to developers and users. Describing data ecosystems in terms of their data supply chain provides a mechanism to identify data sources and the assets which contribute to the development of the data components, or which are produced as the results of intermediate processes.

A metric which enables systems to be rated by the transparency of their constituent components provides a mechanism for users and user communities to establish policies to compare the transparency of models, and use this as a benchmark for building and supporting their own confidence in adopting and using models developed by third parties, or for fostering the development of standards for good quality documentation for internal projects. Providing transparency on the contributions made to an ML model should be considered to be complementary to efforts to provide explainability [16] on the outputs of AI systems, as transparency provides a means to gain assurance on the origins and builders of the system, augmenting the understanding of the system's behaviour that explainability seeks to provide.

In order to develop a quantifiable metric for the transparency of an ML production pipeline, and provide model users with a means to compare the transparency of different models, we look to the literature on supply chain visibility, particularly the work of Caridi et al [17,18] who developed a methodology for inferring a metric for the visibility of a manufacturing supply chain from the point of view of a focus organisation's position in the supply chain. By assigning the focus organisation viewpoint to the end user or auditor of an ML model we can assess the suitability of using Caridi's method as a basis for developing a method for ranking ML production pipelines in terms of contribution visibility, and are able to present an adaptation of Caridi's method which can be used to provide a quantifiable metric to rate the transparency of data-rich systems.

The remainder of this paper is structured as follows: Section 2 identifies and discusses styles of documentation for data sets and ML models proposed in recent literature, and further considers other mechanisms for identifying components of systems from the software and manufacturing domains, Section 3 provides a description of the supply chain visibility model developed by Caridi et al, and presents our contribution by way of a proposal for modifying the method so that it can be applied to ML production pipelines to give a quantified transparency ranking for models, which is illustrated by example in Section 4. Models from popular repositories are used in Section 5 to evaluate the proposed metric mechanism, and Sections 6 and 7 respectively present conclusions and discuss motivations for further work in this area.

## 2 | DOCUMENTING DATA ASSETS AND ML MODELS

The ability to make a judgement on the transparency of a system is dependant on the documentation that is made available to future developers and users of the system. Recently published research proposes several documentation formats for elements of data ecosystems, as well as for AI systems as a whole.

### 2.1 | Documentation of Datasets

Bender and Friedman's *Data statements for NLP*[19] presents a documentation structure which guides the formulation of a description of a dataset in order to provide context for researchers such that they can "understand how experimental results might generalize, how software might be appropriately deployed, and what biases might be reflected in systems built on the software." In articulating contributions to dataset creation Bender and Friedman recognise human roles in the data generation process, identifying the distinct roles of *annotator*– "Annotators may be crowdworkers or highly trained researchers, sometimes involved in the creation of the annotation guidelines"– and *curator* –"Curators are involved in the selection of which data to include". Bender and Friedman also identify the *speaker*, as a result of their NLP domain, which illustrates that different types of data sources will have different entities needing to be mapped and documented.

The paper's authors propose that Data Statements should be included whenever new datasets are presented in publications, and with every NLP system which is built from a dataset, in order to form a "chronology of system development" - a timeline which should include descriptions of the datasets used for model training, tuning and testing. An argument is made for providing two versions of a Data Statement, one which is detailed and published as a research paper or as part of system documentation, and a second more concise statement which is used when describing systems or experiments which make use of the data and which should be used alongside citation of the long-form statement. Data Statements provide a template for dataset creators to consider the context in which their data is used, such that they can expose motivations of the creation process and constraints in generation and use.

Gebru et al. propose *Datasheets for Datasets*[20] which takes inspiration from industries such as electronics and manufacturing, where all components can be referenced to a datasheet detailing the operating characteristics of the device, any test results, guidance on recommended usage, and other pertinent information for users.

Gebru et al. eschew automation in the creation of the datasheet, and identify the task of manually assembling the datasheet as providing an opportunity for researchers to reflect and perhaps alter how they create, distribute and plan to maintain their datasets.

The proposed format provides an exemplar series of open questions about the dataset, encouraging authors to avoid terse answers and to encourage the provision of rich information about the dataset. The format includes a section on Maintenance, which provides an opportunity for detailing how updates or obsolescence of the dataset will be communicated to secondary users. Sample datasheets are provided for the Labeled Faces in the Wild dataset[21] and Pang and Lee's polarity dataset[22].

*The Dataset Nutrition Label*[23] describes a modular framework for presentation of information about datasets, with proposals for web-based tools for authoring and presentation to dataset users. Based on the well-established Nutrition Facts label scheme[24] from the food industry, and building on efforts to broaden this approach exemplified by the Privacy Nutrition Label[25], the Dataset Nutrition Label is presented as a digital object, showing a series of components or modules designed to present information on different aspects of the dataset to prospective users. Some modules will present non-technical information, whereas others can be highly technical, for example presenting machine-generated statistical information about the data in the dataset. The choice of modules presented in each case should be based on the availability of information, the level of willingness and effort volunteered to document the dataset and cognisance of privacy or confidentiality criteria around proprietary datasets. The intent of the system is to offer a flexible and adaptable framework which can be applied across different domains and data types, with an extensible collection of interactive qualitative and quantitative modules displaying their outputs in a standardised format, which is illustrated through a prototype label generated for the Dollars for Docs[26] dataset.

### 2.2 | Documentation of Models and AI Systems

Mitchell et al.'s *Model Cards for Model Reporting*[27] are complementary to Datasheets for Datasets[20]. Model Cards are intended to be short records providing details of the motivations for model development, recommendations for use and quantitative results of evaluation in a document of a page or two in length. The paper's authors see a key role of the Model Card is to provide a documentation framework to promote a standardisation of "ethical practice and reporting" and to allow stakeholders to compare models "along the axes of ethical, inclusive, and fair considerations." As such, they advocate that model performance metrics included in the Model Card are broken down by "individual cultural, demographic, or phenotypic groups, domain-relevant conditions, and intersectional analysis combining two (or more) groups and conditions" along with the reasoning and motivations for choosing such metrics and groupings. By providing information and metrics that capture and highlight bias, fairness and inclusion aspects of a shared ML model's performance and thereby alerting model users to any potential pitfalls it is intended to aid in the mitigation of any negative effects of model deployment. The paper illustrates Model Cards with two examples, a smiling detection model trained on the CelebA dataset[28] and a public toxicity detection model, based on Perspective API's TOXICITY classifier[29].

Highlighting the intended use cases for a model and quantifying the limitations and biases in the model as it is used outside of its design constraints brings potential benefit to a number of different stakeholders. The paper's authors identify a benefit of Model Cards in assisting with the process of curating models for use by less technical users and developers, as they will provide insight into how the model can be combined with other models, along with rules and constraints. In summary, Mitchell et al. argue that researchers sharing models may be incentivised to produce accompanying Model Cards through the benefits they bring to potential users in becoming better informed on the suitability of particular models for particular use cases, as a means to compare and contrast different models.

The structure for a Model Card is not rigidly defined at this stage, but it is recommended that it includes a section on the background to the model's development, detailing as far as is possible information about the model's training and evaluation (and including reference to any Data Sheets or other documentation about the data sets used). The intended use of the model should be declared and described, such that readers can readily understand what the model has been designed to do, and in what contexts, as well as what it should not be used for. This information will also serve to frame the quantitative analysis that it is recommended is included in model cards, as previously discussed.

Mitchell et al. reflect that Model Cards are just one approach to increasing transparency between developers, researchers, and stakeholders of machine learning models and systems, and that they are designed to be flexible, such that they can be used for a wide range of ML models and across different use cases. With this broad application environment in mind, they feel it is unlikely that a standardised form of Model Card will be developed in the near-term, and that documentation in this form is just one tool in providing increased transparency around ML models. As with all efforts to improve transparency through authored documentation, Mitchell et al. identify that the usefulness and accuracy of a Model Card relies on the integrity of the creator of the card itself.

Hind et al. present *FactSheets*[30] which are based on a safety document called the Supplier's Declaration of Conformity (SDoC), which is used in industrial sectors including telecommunications and transportation. The SDoC is usually a voluntary document, which is developed and maintained by component or product suppliers, to provide written assurance of adherence to specified requirements.

The FactSheet document is designed to consider AI systems as a whole, rather than as individual components of a system and as such a FactSheet would be the type of document which might be delivered by a researcher with an ML model or an API for developers to integrate and would form the basis for a review of the visibility of the systems contributing components and overall transparency. As described by Hind et al, AI services are presented as API's to developers, and are potentially an amalgam of many models trained on a variety of datasets. As such, developers are unlikely to directly use models or datasets, and their interface will be through the API offered by the AI service. The scope of the FactSheet document is to "contain purpose, performance, safety, security, and provenance information to be completed by AI service providers for examination by consumers." and is positioned to provide information to fill an identified gap in expertise between the data scientists who created the AI service and application developers or scientists who consume the service in order to provide applications for end-users.

FactSheets are designed to cover aspects of an AI service pertaining to its development and its use, detailing how the service was created and trained, which scenarios it was tested in and how it should respond to untried scenarios, with the intent of grounding service use into particular usage domains. The document will also encourage coverage of security, in particular robustness of the service to adversarial attack, and safety. The holistic approach is proposed to enable a "functional perspective" to be taken on the overall service and tests being conducted and documented for aspects of performance, safety and security which are not relevant for the individual components of the system, such as accuracy, explainability, and adversarial robustness.

Hind et al. present exemplar FactSheets for two hypothetical situations, a fingerprint verification service and a service identifying trending topics in social media. The samples provide rich coverage of information about the hypothetical services and the factors which led to their creation, the contributing assets and documentation of the testing, safety and security of the services. The FactSheets encourage the attachment of datasheets or supporting information about assets, which would provide a route back to the source information for developers trying to gain further insight and visibility into the makeup of AI systems that they are trying to assess in order to make judgements on transparency.

## 2.3 | Documentation Standards from Other Industries

Outside of the AI domain, the Open Data Institute[31] has described and illustrated efforts to map the contributors and stakeholders in situations where multiple parties cooperate over the provision and use of data sources in scenarios such as Transport for London's provision of transport data[32] via an open API. These data ecosystem mappings provide a visual overview of the stakeholders in the systems, along with connections showing the data flows and value flows[33] between the parties. If one considers that an AI system or an ML model is an output asset of such a data ecosystem, wherein multiple actors cooperate to add value to data sets, then a visual data ecosystem map could provide a suitable vehicle for providing context to the parties who have made a contribution to the AI system and be useful in augmenting textual documentation[34] about the details of contributions to end-users and developers. Presenting a visual overview of the entire AI system can readily identify the stakeholders and contributors, such that documentation for each contributing entity can be understood in the context of its role in the system as a whole.

The familiar model of a supply chain, and its record of the sub-assemblies used in the construction of a physical product provides a convenient model for recording the contributions made to the development of a complex digital product. Jansen-Vullers, van Dorp, and Beulens[35] and van Dorp[36] discuss the composition of physical products in terms of a Bill of Materials (BoM) which is the list of types of component needed to make a finished item of a certain type. In manufacturing, a BoM would specify which sub-assemblies were used to produce a product. Furthermore, a BoM can be multi-level, wherein components can be used to create sub-assemblies which are subsequently used in several different product types.

The authors' previous work[14] has discussed how ML models and other aggregated data assets can be broken into component parts, supporting artefacts and human work and itemised and mapped in a BoM to recognise and provide traceability on contributions to the asset. In the software engineering community, the increasing use of software components from multiple suppliers, including open source components, in the development of software systems has led to recommendations for a Cyber Bill of Materials (CBOM) or Software Bill of Materials (SBOM) which documents the supply chain to be created and maintained and delivered alongside software systems or devices employing software for their operations[37]. In such a way, outdated sub-components of the system can be readily identified, which is particularly important when vulnerabilities have been found and fixed in updated versions of the module[38]. Tools including CycloneDX[†], SPDX[‡], and SWID[§] are defining formats for identifying and tracking such sub-components. The SBOM also provides a placeholder for supplementary but important information such as licensing conditions for individual software components and libraries used, so that consuming parties can take measures to ensure that they are not breaching license terms.

Providing well structured and comprehensive documentation with shared data assets and ML products provides an opportunity for researchers to give potential users an opportunity to understand the supply chain of the asset, such that they can build a picture of the contributions made to the asset and make an informed judgement as to its qualities and suitability.

## 3 | QUANTIFYING SUPPLY CHAIN VISIBILITY

The study of visibility in manufacturing supply chains is an established field, which Parry et al.[39] summarise, and identify three constructs for characterising visibility, namely "the exchange or sharing of information", "the properties of information exchanged" and "the usefulness of information exchange or a capability to act on information exchange".

A discussion of visibility on published digital assets is taken up by McConaghy et al.[40], who make the case that the unidirectional hyper-linked nature of the world wide web leaves a lack of opportunity for dialogue between the publisher and consumer of digital assets, such that the consumer is party only to the information made available by the asset owner at publication time. If the publisher only shares a limited amount of information about the asset, then information asymmetry occurs almost by default, with the consumer of the information unaware of unreported information, such as any usage rights associated with the asset. McConaghy et al. assert that "information availability helps both initiate and inform action, thus impacting upon an individual's decision making process".

---

[†]https://cyclonedx.org
[‡]https://spdx.org"
[§]https://www.iso.org/standard/65666.html

In seeking to provide a metric for levels of transparency on the ML production supply chain for a particular model, the emphasis is put on providing users or auditors of the output products of the systems with the capability to act upon information about the contributions to the supply chain, with the metric for visibility providing a means of determining the extent to which relevant and useful information is available to make decisions and judgements about the suitability of the system. As such, it can be argued that a production pipeline with a high-value visibility metric, relative to the scale, will provide a good level of useful information about its data sources, contributions and processes - the "dimensions of data transparency" proposed by Bertino, et al.[15] - and a system with a low value visibility metric will provide a minimal amount of information, or information of low quality.

Caridi et al. have proposed[17] and later refined[18] a method for providing a quantifiable measure for the overall visibility of a supply chain from the point of view of a focus organisation, wherein supply chain managers make semi-quantitative judgements on information available at each node in their network according to three scales - the quantity of exchanged information, the quality or accuracy of the information, and the freshness of the information. Information in the following categories are considered: transactions and events, status information, master data and operational plans, with judgements made for each of the four information categories for each node in the supply chain. Scores are awarded from 1 (lowest) to 4 (highest), when evaluated against a qualitative scale for each information category.

Caridi produces twelve judgements for each node, which are numbers from 1 to 4, according to a scale provided for each information category and each measure. Judgements assigned for information freshness and accuracy are combined to give an information quality index for each node, which is then combined with the quantity judgement to give an overall visibility rating for the node. Caridi's model then weights each node based on its closeness to the focus organisation, and its overall impact on the system, and the weighted nodes are combined to give an overall rating for the visibility of the supply chain as a number in the range from 1, for systems with the least visibility, to 4 for the highest.

Caridi's method has previously been used outside its intended domain by Vlietland and van Vliet[41], who adapted the model to quantify visibility of performance requirements for incident handling in IT departments. Vlietland and van Vliet used Caridi's dimensions of accuracy and freshness of information, but did not use quantity. Accuracy was used as a measure of the required and delivered performance for each node, and freshness as the timeliness of the information.

| Score | Quantity | Freshness | Accuracy |
| --- | --- | --- | --- |
| 1 | Sparse or insufficient information | Never updated | Demonstrably inaccurate |
| 2 | Some information missing | Out-of-date | Believed to be inaccurate |
| 3 | Sufficient to gain confidence | Updated when changed | Believed to be accurate |
| 4 | Sufficient to validate | Real-time validation | Evidenced and verifiable |

**TABLE 1** A scale to judge documentation on each contribution to ML models or data assets

To determine a visibility ranking for a data supply chain for a machine learning model or other produced data asset, it is proposed to use the Master Data category and adopt the same scales used in Caridi's model, but define them in a context suited to a data domain[42]. Initially a qualitative description will be written (Table 1) to allow the scoring from 1 to 4 depending on an assessment made for each node based on interpretation and judgements made on the information in the documentation supplied with the model or data asset. When assigning scores for each of the scales, determination of the ratings should be made in regards to data sources, datasets and human participants contributing to the generation of the model or data asset, taking into consideration all information that is shared with the user or auditor of the asset. Future research will attempt to identify a set of objective measures that will be capable of being mechanised to replace the subjective elements of the ranking. A Bill of Materials document supplied or made available with the model, as proposed[43], would be a suitable vehicle for making such information about the contributors available to model users, as it facilitates both the identification of significant contributions to the system (the nodes) as well as providing a means to identify supporting information and artifacts. Other proposals for documentation of ML systems, such as the Model Card proposal from Mitchell et al.[27] or a Supplier's Declaration of Conformity as suggested by Hind, et al[30], could also be used as the basis for a framework which delivers information from which the visibility judgements could be made.

| Judgement Criteria | Judgement |
|---|---|
| Quantity | $j_q$ |
| Accuracy | $j_a$ |
| Freshness | $j_f$ |

**TABLE 2** Notation for node ranking judgements

By making judgements for each contributing node, $k$, in the data supply chain against a set of defined criteria, as exemplified in table 2, it can be determined that the node's Visibility Quantity Index is:

$$VISQuantity_k = j_q$$

and for each node, $k$, the Visibility Quality Index is:

$$VISQuality_k = \sqrt{j_a \times j_f}$$

with the node's Visibility Index being:

$$VIS_k = \sqrt{VISQuantity_k \times VISQuality_k}$$

In Caridi's model, each node is weighted according to its impact on the system, such that the overall visibility index *VIS* of the supply chain is given:

$$VIS = \sum_{k=1}^{M}(VIS_k \times W_k)$$

In determining a transparency ranking for data supply chains, it is proposed to initially assign an equal weighting $W_k$ to each contributing node, with future research considering the impact of assigning different weightings for individual contributors in different system configurations.

As such, the visibility index for a data asset or ML model resulting from a pipeline with *M* contributing nodes can be determined as:

$$VIS = \sum_{k=1}^{M} \frac{VIS_k}{M}$$

Where *VIS* will be a number in the range 1 at the low end, to 4 at the high end for systems with very high levels of visibility on contributions towards the output.

## 4 | APPLICATIONS OF THE METHOD

In order to assess the viability of using the proposed variant of Caridi's method to provide a quantified measure of the transparency of an ML pipeline, an example scenario illustrating the production of a simple machine learning model is rated against the information ranking criteria suggested in Table 1.

The example pipeline (Figure 1) illustrates a simple ML model training scenario, and contains a data source (DS), which has been labeled by a curator (H1) to produce a labeled dataset (LD). An AI engineer (H2) uses the labeled dataset (LD) to train and test an ML model (M), which is uploaded to a model zoo so it can be used by third parties. Note that we only need to consider the visibility of contributions from leaf nodes in the system, DS, H1 and H2. LD is a created asset and as such its transparency rank is determined by its contributing nodes.

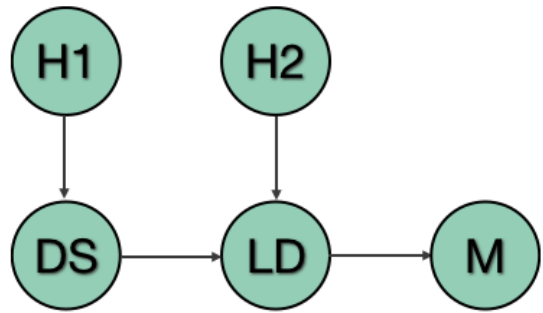

**FIGURE 1** Production pipeline for a simple ML training process

This pipeline can be applied to multiple machine learning paradigms that are currently in widespread use, with all aspects of the model's contributing factors attributed to one of the 3 nodes: DS, H1, H2. There are paradigm-agnostic attributions that can be made, for example, the decision for how to split the labelled dataset (LD) into training and testing partitions can only be made by the the AI engineer (H2). There can be no technical restrictions imposed on the labelled dataset (LD) by the curator (H1), only guidelines for performance bench-marking of the resulting model (M).

More fine-grained attributions need to be made depending on the ML paradigm chosen, with the evaluator requiring competency in the type of ML used. If M is a decision tree[44] then H2 will have made some decision about which algorithm to use (CART or ID3), the splits and tree depth used, as well as other parameters that are used in the final model. Another example is if M is a deep neural network; meta-information including the network architecture, loss function, and data transforms applied to LD are all decisions made by H2. In both cases, these attributions need to be assessed and reflected in their visibility score by a reviewer who is familiar with the technical area.

To understand the impact of different levels of access to supporting information about the contributions made to the generation of the model on its transparency rating a set of scenarios are explored, these range from the point of view of the organisation which produced the model (where one would anticipate visibility would be high) to models shared with varying quality levels of supporting documentation.

## 4.1 | First Party Model Usage

The first scenario considers a hypothetical model which has been generated recently, and is used in the same laboratory in which it was developed, such that the users of the model can be assumed to have or to be able to gain access to complete information on the origins of the training dataset and the researcher's contributions to the system.

| Node | Quantity | Freshness | Accuracy | *VISQual* | VIS |
|---|---|---|---|---|---|
| DS | 3 | 3 | 3 | 3 | 3 |
| H1 | 4 | 4 | 4 | 4 | 4 |
| H2 | 4 | 4 | 4 | 4 | 4 |
| Overall *VIS* for model | | | | | **3.67** |

**TABLE 3** Scores assigned to first party model usage scenario

In such a case, it might be reasonable to apply high ranking scores to each of the judgements for each contribution to the system, with some variation likely where data sources were externally sourced, or if the researcher producing the model had to be located and interviewed. Thus, from the point of view of an ML model user with good and recent knowledge of the training data, the data curator and the researcher responsible for training the model, it could be envisaged that the transparency rating for the model would have a high score compared to the maximum 4.0, as depicted by Table 3.

| Node | Quantity | Freshness | Accuracy | *VISQual* | *VIS* |
|---|---|---|---|---|---|
| DS | 3 | 2 | 3 | 2.45 | 2.71 |
| H1 | 3 | 3 | 3 | 3 | 3 |
| H2 | 3 | 3 | 3 | 3 | 3 |
| Overall *VIS* for model | | | | | **2.90** |

**TABLE 4** Judgements on a first party model after a few years

As time passes, an ML model might remain in use in an organisation but the staff responsible for the development of the model could move on to different projects or depart to other organisations. It may still be necessary to assess the model's transparency from time-to-time, to determine whether it is still suitable for use according to legal or ethical standards of the day, or if regulators or auditors require inspection. In re-evaluating the information supplied with the model through its documentation, the freshness will naturally have degraded (unless processes are in place to maintain this information and are followed adequately), and it might be that it is no longer possible to have the same confidence in the accuracy of the information supplied about the data source or personal regard for the staff involved in the curation of the data or the model training. Further, a new witness may find gaps in the information available about the contributors, such that they are unable to validate assertions made. This re-evaluation of the ranking criteria might lead to new judgement scores, as in Table 4, which shows a degradation in the transparency ranking of the model, as clarity and confidence in the assertions about the source data and human contributors diminishes over time, reducing the score from the initial determination to a more pragmatic 2.9 rating.

## 4.2 | Third Party Model Usage

Increased sharing of ML models and generated data assets through science gateways, model zoos and commercial marketplaces mean that the users of the model or the asset may have very little connection with the original researchers, and minimal insight into how the asset was made, and in particular the qualities and qualifications of the data sources and the scientists or engineers involved. There will be situations such as inspections, audits or even legal challenges where users are required to demonstrate that their systems are suitable for use and meet the necessary regulations for their industry or domain. It is appropriate, therefore, that users of ML models and data assets are able to make an informed decision on the suitability of the systems they use by being guided to form a judgement on the degree of transparency that they have on the production of the assets, such that if necessary they can trace the assets or contributors and demonstrate that they are still suitable for use. An example might be a model that was developed and published in an ML community in 2018 and is still in use within an organisation in 2025, by which time it forms a central part of a research pipeline. In performing due diligence, an auditor may seek to understand the origins of this model, such that they can gain assurance that it was produced using trustworthy data, and that the researchers who produced it were suitably trained or qualified. The assertion being that a system with a high transparency metric will enable such checks and assurances to be efficiently made, whereas a system with low transparency may prove impossible to validate, leading to uncertainty and potentially inconvenience or high cost as the model is replaced.

A common use-case for third party models is to employ transfer learning[45] in order to develop strong predictive performance by using a previously trained model's hyper-parameters as a basis for learning (referred to as the basis model: $M_{basis}$). This is typically used where an organisation's computational resources or data availability are constrained such that a well performing model cannot be feasibly developed using the resources available to the AI engineer (H2). In this use-case the resulting model (M) needs to be evaluated after the basis model ($M_{basis}$) has already been evaluated as this gives the reviewer full visibility of the entire data supply chain. We consider the data that $M_{basis}$ was trained on as being contained as part of H2's contributions, and is subject to it's own self-contained review. Finally, depending on the development process either H1 or H2 needs to ensure that the labelled dataset (LD) is suitable for use when using $M_{basis}$ as the starting point for training the final model (M). This suitability is determined by how well aligned the intended tasks for $M_{basis}$ and M are.

The scenario presented here considers a model which has sparse supporting documentation, such that it is very difficult for a third party to gain insight on the individual contributions made towards its production. Judgements for such a system are shown in Table 5, and are based on the criteria previously shown in Table 1.

| Node | Quantity | Freshness | Accuracy | *VISQual* | VIS |
|---|---|---|---|---|---|
| DS | 1 | 1 | 3 | 1.73 | 1.31 |
| H1 | 1 | 1 | 3 | 1.73 | 1.31 |
| H2 | 1 | 1 | 3 | 1.73 | 1.31 |
| Overall *VIS* for model | | | | | **1.31** |

**TABLE 5** Judgements on a poorly documented third party model

Accordingly, the sparsely documented system gives rise to a very low transparency score, which should serve as a warning to organisations not to allow it to become a mission critical asset.

In the second instance of the scenario, consider a professionally produced and packaged ML model which comes complete with full supporting information, and perhaps a contact email address or an accessible API to facilitate real-time queries to be made on the status of the contributing assets. Such rich documentation would provide assurance of the qualifications of the contributing staff, and information about the data sources.

| Node | Quantity | Freshness | Accuracy | *VISQual* | VIS |
|---|---|---|---|---|---|
| DS | 4 | 3 | 3 | 3 | 3.46 |
| H1 | 4 | 3 | 3 | 3 | 3.46 |
| H2 | 4 | 3 | 3 | 3 | 3.46 |
| Overall *VIS* for model | | | | | **3.46** |

**TABLE 6** Judgements on a well documented third party model

As Table 6 demonstrates, a well-documented model from a third party source can score well for transparency, and over a number of years may perform better than a poorly documented in-house system, Table 4, as locally available knowledge degrades over time, which is illustrated clearly in the transparency ranking for the system. Scores could increase further with demonstrable evidence of qualifications, but this would need to be tempered by the need to protect staff privacy and commercial secrets.

## 5 | EVALUATING THE METRIC WITH MODELS FROM SHARED REPOSITORIES

In this section we will evaluate the descriptive ability of the proposed metric on models retrieved from two model zoo repositories that are commonly used by the research community, namely PyTorch Hub[46] and ModelHub[47]. Although the final evaluation comes down to the engineers and data involved, different gateways and model zoo platforms may help or hinder this process in the implementation of their hosting format.

### 5.1 | ModelHub

ModelHub[47] is a model zoo with a focus on ease of access and rapid deployment of deep neural network models. The zoo's online interface allows for inspection of the model's description (e.g. intended application, training procedure, data domain etc.), as well as a visualisation of the model architecture and a live demonstration based on sample data. Once a model has been downloaded, the model zoo provides a sand-boxed environment based on Docker[48] to interact with it. This reproducible environment ensures that the model that is being deployed is identical to the one being offered, with no possibility for the end user to manipulate the model's internal parameters. This however means that the model's internal processing is completely abstracted from the end user. There is the possibility for inspecting the source code for the model if it has been made available to the model zoo, but this is at the discretion of the developer who uploaded the model to the repository.

For this evaluation, the proposed **VISQual** and **VIS** metrics were applied to a pre-trained GoogleNet model[49] that is available via ModelHub [¶]. GoogleNet is a convolutional deep neural network model trained on the ImageNet dataset[50] for the task of general image classification. The following attributions are made for this instance of the model:

- **DS**: A collection of real world images sourced from internet search engine results
- **H1**: Deng et al.[50] curated ImageNet (**LD**) from the real world images **DS**
- **H2**: The developers and maintainers of ModelHub[47]

The model that is made available via the ModelHub API (**M**) has been sourced by **H2** from the Open Neural Network Exchange project[51], which hosts a trained replication of the model described in Szegedy et al.[49] [#]. We find that the technical documentation in the form of the source code repository accompanied with the original paper is sufficient for a high quantity of information score for **H2**, however, at time of writing the URL provided as the provenance for the model does not resolve, [‖] leading to missing information about the source of the model. This means that the freshness of information is out of date, however the quantity of information available does offset this factor as it means that one could feasibly compare the original model and the model that is provided via the model zoo **M**. The documentation for the curation procedure of the ImageNet dataset[50] is sufficient to validate and gain confidence about the overall distribution and content of the labelled dataset (**LD**), resulting in good visibility scores for **H1**. However the nature of the automated process for gathering the original data (**DS**) means that it is impossible to inspect and validate every contribution of data made to the dataset. Table 7 shows the scores assigned to GoogleNet after due consideration.

| Node | Quantity | Freshness | Accuracy | *VISQual* | *VIS* |
|---|---|---|---|---|---|
| DS | 1 | 1 | 3 | 1.73 | 1.31 |
| H1 | 3 | 3 | 4 | 3.46 | 3.22 |
| H2 | 4 | 2 | 3 | 2.45 | 3.13 |
| Overall *VIS* for model | | | | | **2.55** |

**TABLE 7** Scores assigned to GoogleNet model from ModelHub model zoo

## 5.2 | PyTorch Hub

PyTorch[46] Hub is a model zoo that provides access to existing PyTorch models by wrapping the code required to download and initialise deep neural network models hosted on open source git repositories. Each model's description page provides direct access to the model's source code, as well as a quick summary of intended use-cases. As PyTorch Hub doesn't have a curation procedure for hosting models, it is the responsibility of the developers of the submitted models to ensure that the documentation in the model's repository is of good quality. For the evaluation of PyTorch Hub the **VISQual** and **VIS** metrics we applied to the DistilBERT[52] [**] model available on the model zoo. This model was chosen as it highlights a complex scenario in which the training data comes from multiple sources, as well as how to address the visibility of contributions for models that use transfer learning. BERT[53] is a language comprehension model commonly used in natural language processing tasks that achieves near state of the art performance, but requires substantial computational resources to run. DistilBERT has been designed to have smaller computational requirements but with similar performance when compared to the original model. This is achieved via the process of model distillation[54], where a student model is trained with the goal of replicating the output of the teacher model. For this model the following attributions were made:

---

[¶] Available at http://app.modelhub.ai/.
[#] Available at https://github.com/BVLC/caffe/tree/master/models/bvlc_googlenet
[‖] https://github.com/onnx/models/tree/master/bvlc_googlenet
[**] Available at https://github.com/huggingface/transformers/tree/master/examples/distillation

- **DS₁**: The Toronto Book Corpus [55].
- **DS₂**: English Wikipedia.
- **H1**: Devlin et al. [53] proposed the original BERT model as well as curating the dataset combination **LD** from **DS₁** and DS₂.
- **H2**: Sanh et al. [52] used the original BERT model as a basis to train DistilBERT **M**.

| Node | Quantity | Freshness | Accuracy | *VISQual* | *VIS* |
|---|---|---|---|---|---|
| DS₁ | 3 | 2 | 3 | 2.45 | 2.73 |
| DS₂ | 3 | 1 | 3 | 1.73 | 2.28 |
| H1 | 4 | 3 | 3 | 3 | 3.46 |
| H2 | 4 | 4 | 4 | 4 | 4 |
| Overall *VIS* for model | | | | | 3.12 |

**TABLE 8** Scores assigned to DistilBERT hosted on PyTorch Hub model zoo

Although some documentation of the contributing data (**DS₁**, **DS₂**) is sparse[††], the documentation and availability of the model's source code results in an overall good **VIS** score. The documentation from the authors is provided in multiple forms: the source code, manuscript, as well as active community via the software repository. After consideration against the evaluation criteria, Table 8 shows the scores assigned.

## 5.3 | Discussion

Applying the proposed **VIS** metric to two deep neural network models available from popular model zoos has shown that the metric can highlight gaps in documentation in the presentation of deep neural network models. For example, ImageNet's data source in 5.1 has an overall low visibility metric due to the procedure of collection, however, the quality of documentation on how the dataset was curated combined with the extensive documentation about the model brings the overall **VIS** to a reasonable level. When comparing this to the data sources in 5.2, the documentation associated with the two respective sources as well as the intended usage is sufficient enough to gain confidence that the data source is providing the right contributions. Both models from ModelHub (5.1) and PyTorch Hub (5.2) have good overall visibility scores due to the availability and quality of documentation around the training and architecture of the models.

The metric provides a framework that does not overly penalise lack of proper validation or stale information if there is sufficient relevant documentation overall. Models that are hosted and used extensively by the research community share a similar profile to that found in 5.1, which is that the original trained model has been reformatted to be compatible with the tools that are in common use. The code base may be completely different, but the underlying model is exactly the same. As long as the conversion procedure is trusted to be correct (such as the ONNX project [51]) or it allows like-for-like comparisons, then the quality of the original documentation still stands. The model evaluated in 5.2 benefits from being a recent contribution, and as such the authors are very active in the related community providing additional documentation. We can see that the freshness is a strong indicator for the **VISQual** metric, however the quantity of informational documentation can outweigh this for the overall **VIS** score. A perfect score can be achieved by providing full access to model training and inference code as well as data sourcing procedures, however this is a scenario that in practice is unlikely to be realised. As such, if this metric were to be taken into account when developing and documenting AI systems, then developers should be motivated towards making contributions that are designed to last, by providing quality documentation that is general enough to describe the replication of a model and dataset.

---

[††]For instance, the original release of the Toronto Book Corpus [55] release is no longer publicly available, and there is no documentation on how exactly English Wikipedia was obtained

# 6 | CONCLUSIONS

The ability to assign a quantitative ranking to the transparency of the processes and contributions leading to the development of an ML model or data asset provides a motivation for scientists and researchers to seek insight and make a judgement on the suitability of a model for use in their organisation, and then to be in a good position to monitor their on-going confidence in the suitability of the model over a number of years. The work presented here has demonstrated that guided judgements can be made on the quantity and quality of the supporting information for each contribution in an ML or AI system production pipeline and does impact the transparency rating for the system as a whole.

It has been shown that the use of an adaptation of the method to develop a supply chain visibility metric proposed by Caridi can be of value in determining the level of transparency afforded into a data supply chain and the role of its contributors, such that systems can be evaluated and compared on the basis of their transparency. Rankings resulting from the analysis process have the potential to be used as a mechanism for driving improvements in the design and provision of documentation provided with models and data assets. The use of a well understood quantitative transparency rating will become increasingly important as the distance between AI system developers and system users grows, which is inevitable as models get deployed and used for key processes within organisations, and as they are shared and used by third parties. As such, motivation to provide and maintain accessible, up-to-date and trustworthy documentation supported by machine readable and verifiable evidence of the contributions made to ML model or data asset development will become increasingly important, and will serve to assure users and other stakeholders that the datasets used for training the model have not been discredited, and that the staff used in data creation, curation or engineering were appropriately trained and qualified for their tasks.

# 7 | FURTHER WORK

The strength of the subjective criteria used to assign rankings to the contributors of the asset will vary between organisations, and it would be helpful to see further discussions on what these criteria should be, such that consensus and standard terminology can begin to emerge. Ideally, progress will be made towards determining measurable and objective criteria, such that subjective evaluations and judgements and their impact on rankings can be minimised. Whilst software development processes are more mature than AI system processes, the work of the software engineering community towards providing documentation generated or augmented automatically from build processes in the form of the SBOM provides a useful model for a direction that the AI community could be taking, and which research by Schelter at al.[56] in extracting and recording metadata as part of process of model generation can help to facilitate. By automating the compilation and verification of the aspects of documentation that can be readily automated, manual documentation effort can be invested towards providing a higher quality of information where it is most needed. Provision of machine-verifiable information will also facilitate a more rapid and accurate transparency measure to be generated for each component and for the system as a whole, and help to remove subjectivity from judgements. Providing a means to determine the freshness of information would appear to be one of the more straight forward aspects of the system to automate, and should be able to leverage work being done in the open source software community, for example research by Coelho et al.[57] which aims to identify software repositories which have been abandoned. By applying similar techniques to datasets, shared models and their documentation, it should be possible to systematically arrive at freshness values, or at the least to indicate to stakeholders that the assets they are using appear to be no longer maintained.

Ultimately it would be beneficial if the results of the transparency metric evaluation could be shown to users at the point of selection of a model, through integration with the workflow of a science gateway or model zoo platform. Whether this is achieved through manual curation by the repository's operators, or a score is aggregated by crowdsourcing from the platforms's users, it would be a useful comparator and quality indicator of models under consideration by scientists. As such, cooperation with a science gateway operator would enable research into the ideal presentation of a transparency metric and its value to the scientific community to be conducted.

A utility of the proposed metric is to indicate to users how well they will be able to assess and maintain ongoing insight into systems once in deployment, and after the passing of time. In order to provide the means to maintain and improve transparency rankings on systems after many years, and as they extend in use away from the originating organisation, further work is to be conducted in finding ways to provide measurable and verifiable evidence of contributions. This could include providing mechanisms to certify the qualifications of staff, or providing systems to notify model users of any issues around the legitimacy of data - for example, if the data source or labelled dataset is later found to be corrupt or have contained unexpected bias which would affect the legitimacy of the model once it is in use.

The authors have begun to investigate[58] the use of linked data documents and mechanisms for decentralised identifiers (DIDs)[59] and cryptographically verifiable credentials[60] or attestations as an approach to enabling automated checks on claims and credentials to be made about attributes of digital assets and entities, whilst protecting the privacy of both the subject and the verifying party, particularly in environments where there is no direct relationship between the organisations. Further work will integrate these techniques with a BoM-style document detailing the contents of an aggregated data asset or a model in a way which can provide machine-readable cryptographic verification of the provenance of data sources and individuals working on data sources and algorithms, in order to provide increasing automation of ongoing visibility and confidence in the veracity of data rich assets.

## ACKNOWLEDGMENTS

This research was sponsored by the U.S. Army Research Laboratory and the UK Ministry of Defence under Agreement Number W911NF-16-3-0001. The views and conclusions contained in this document are those of the authors and should not be interpreted as representing the official policies, either expressed or implied, of the U.S. Army Research Laboratory, the U.S. Government, the UK Ministry of Defence or the UK Government. The U.S. and UK Governments are authorized to reproduce and distribute reprints for Government purposes notwithstanding any copyright notation hereon.